\documentclass[11pt]{article}

\usepackage[final]{acl}

\usepackage{times}
\usepackage{latexsym}

\usepackage[T1]{fontenc}

\usepackage[utf8]{inputenc}

\usepackage{microtype}

\usepackage{inconsolata}

\usepackage{graphicx}

\usepackage{amsmath,amsfonts,bm}

\def\eqref#1{equation~\ref{#1}}

\def\1{\bm{1}}

\def\vs{{\bm{s}}}

\def\mC{{\bm{C}}}

\def\mQ{{\bm{Q}}}

\DeclareMathAlphabet{\mathsfit}{\encodingdefault}{\sfdefault}{m}{sl}
\SetMathAlphabet{\mathsfit}{bold}{\encodingdefault}{\sfdefault}{bx}{n}

\usepackage{graphicx}
\usepackage{hyperref}
\usepackage{url}
\usepackage{algorithm}
\usepackage{algorithmic} 
\usepackage{amsthm}
\usepackage{wrapfig}
\usepackage{threeparttable}
\usepackage{booktabs}
\usepackage[table]{xcolor}
\usepackage{multirow}
\usepackage{makecell}
\usepackage{cleveref}
\usepackage{appendix}
\newcommand{\methodname}{\texttt{MATCH}}

\usepackage[slantedGreek]{newtxmath}

\def\Snospace~{\S{}}

\let\orgautoref\autoref
\providecommand{\Autorefs}
        {\def\figureautorefname{Figs.}%
         \def\subfigureautorefname{Figs.}%
         \def\Itemautorefname{Items}%
         \def\tableautorefname{Tables}%
         \orgautoref}

\renewcommand{\autoref}
        {\def\figureautorefname{Fig.}%
         \def\subfigureautorefname{Fig.}%
         \def\sectionautorefname{\Snospace}%
         \def\subsectionautorefname{\Snospace}%
         \def\subsubsectionautorefname{Snospace}%
         \def\appendixautorefname{Appendix}
         \def\Itemautorefname{item}%
         \def\tableautorefname{Table}%
         \orgautoref}
         
\usepackage{etoolbox}
\makeatletter
\patchcmd{\hyper@makecurrent}{%
    \ifx\Hy@param\Hy@chapterstring
        \let\Hy@param\Hy@chapapp
    \fi
}{%
    \iftoggle{inappendix}{
        \@checkappendixparam{chapter}%
        \@checkappendixparam{section}%
        \@checkappendixparam{subsection}%
        \@checkappendixparam{subsubsection}%
        \@checkappendixparam{paragraph}%
        \@checkappendixparam{subparagraph}%
    }{}%
}{}{\errmessage{failed to patch}}

\newcommand*{\@checkappendixparam}[1]{%
    \def\@checkappendixparamtmp{#1}%
    \ifx\Hy@param\@checkappendixparamtmp
        \let\Hy@param\Hy@appendixstring
    \fi
}
\makeatletter

\newtoggle{inappendix}
\togglefalse{inappendix}

\apptocmd{\appendix}{\toggletrue{inappendix}}{}{\errmessage{failed to patch}}
\apptocmd{\subappendices}{\toggletrue{inappendix}}{}{\errmessage{failed to patch}}

\title{\includegraphics[height=15pt]{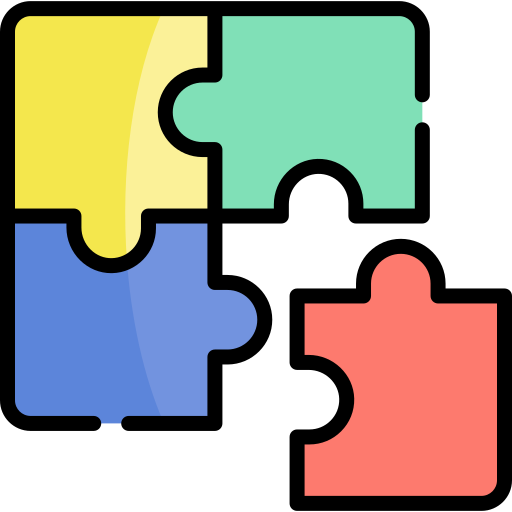} \methodname: Modulating Attention via In‑Context Retrieval for Long‑Context Transformers}

\author{\textbf{Linrui Ma}$^\spadesuit$\thanks{Equal Contribution.}
    \ \ 
    \textbf{Chun Hei Lo}$^\spadesuit$\footnotemark[1] 
    \ \ 
    \textbf{Xinyu Wang}$^\clubsuit$\footnotemark[1]
    \ \ 
    \textbf{Peng Lu}$^\heartsuit$
    \ \
    \textbf{Xihao Yuan}$^\diamondsuit$
    \\ 
    \textbf{Hanting Chen}$^\diamondsuit$
    \ \ 
\textbf{Kai Han}$^\diamondsuit$
    \ \ 
\textbf{Xinghao Chen}$^\diamondsuit$
    \ \ 
\textbf{Chengjun Zhan}$^\diamondsuit$
    \ \ 
\textbf{Hanlin Xu}$^\diamondsuit$
    \\ 
\textbf{Yichun Yin}$^\diamondsuit$
    \ \ 
\textbf{Lifeng Shang}$^\diamondsuit$
    \ \ 
\textbf{Feng Wen}$^\spadesuit$
    \ \ 
    \textbf{Boxing Chen}$^\spadesuit$
    \ \ 
    \textbf{Yufei Cui}$^\spadesuit$\thanks{Corresponding author: \href{mailto:yufei.cui@huawei.com}{yufei.cui@huawei.com}}
    \\ \\
    \textsuperscript{$\spadesuit$}Huawei Canada \ \textsuperscript{$\clubsuit$}McGill University \ \textsuperscript{$\heartsuit$}Universit\'{e} de Montr\'{e}al \ 
    \textsuperscript{$\diamondsuit$}Huawei
}

\begin{document}
\maketitle
\begin{abstract}
The quadratic computational cost of traditional attention mechanisms poses a major bottleneck to the scalability and practical deployment of large language models (LLMs), particularly in long-context scenarios. To improve efficiency, existing approaches often enforce rigid structural constraints such as local attention windows. However, these strategies typically lead to substantial performance degradation on tasks requiring precise long-range recall. In this work, we propose \methodname, a scalable and efficient framework that augments sparsified attention mechanisms with dynamically integrated in-context information through an efficient retrieval system. Empirical results show that \methodname~significantly improves the performance of sparse-attention models on both synthetic and real-world natural-language tasks. These findings highlight the versatility of \methodname~as a general approach for enhancing in-context retrieval capabilities while maintaining the efficiency benefits of sparse attention architectures.

\end{abstract}

\section{Introduction}

Large language models (LLMs) have demonstrated remarkable capabilities across a wide range of natural language processing tasks~\citep{gpt4, llama3, phi3, deepseek, qwen3}. Yet their scalability is fundamentally constrained by the \emph{quadratic} computational complexity of the self-attention mechanism. This bottleneck becomes particularly acute when the context length is extended, as both memory and computation costs grow rapidly with the sequence length~\citep{page-attention, Mell}. Consequently, practical deployments of LLMs are often forced to operate with truncated contexts, limiting their ability to fully exploit long-range dependencies in data. Numerous methods were developed in the hope of retaining inference efficiency and competitive performance as full attention while having light-weight memory consumption~\citep{H2O, scissorhands, attntion-sink, lm-infinite, snapkv}. 

However, it is hard to reduce the memory footprint of the KV cache without sacrificing performance on challenging tasks. A common line of work seeks to mitigate this limitation by replacing full attention with sparse attention mechanisms (e.g., sliding window attention~\citep{SWA}) or generalized linear attention (Mamba~\citep{mamba, mamba2}) to reduce computational and memory overhead from quadratic to linear or sub-quadratic complexity. While effective in enhancing computational efficiency, their local window patterns or fixed-size recurrent state significantly constrain the model’s capacity to retrieve and integrate pertinent information across distant positions within a sequence~\citep{zoology_mqar, swa_not_far}. Consequently, such limitations often lead to diminished performances on tasks that demand precise recall of contextual details.

In this work, we address this trade-off between efficiency and recall by augmenting sparse-attention LLMs with an \textbf{external retriever} designed to enhance long-range recalling capabilities. This retriever acts as a complementary module that processes the entire context in a computationally efficient manner, encoding salient information and making it accessible to the sparsified LLM. By integrating the retriever’s outputs into the sparse-attention layers, our method preserves the efficiency benefits of sparsity while substantially improving the model’s capacity to retrieve and reason over distant information.

Our contribution can be summarized as follows: First, we introduce a novel architecture that pairs sparse attention with an external retrieval-enhancing encoder and propose a retrieval-enhanced attention. Then, we conducted extensive experiments on both synthetic ICL tasks and real-world long-context benchmarks and demonstrate substantial improvements over baseline sparse-attention models. Finally, we perform a comprehensive analysis of its efficiency, including throughput and memory footprint.

\section{Related Work}
\paragraph{Retrieval-Augmented Language Models.} Retrieval-augmented language models enhance performance on knowledge-intensive tasks by incorporating a dedicated retrieval module that sources relevant textual information from an external knowledge base~\citep{rag, rag1, rag2}. This retrieved content is then combined with the original input and passed to the main model, allowing it to access information beyond its parametric memory~\citep{parameters-kb}.
~\citet{Atlas} conduct a deeper analysis of how different loss functions influence retrieval module performance. Subsequent advancements have focused on refining passage selection~\citep{self-rag, ma2025thinkongraph}, improving resilience to irrelevant or noisy retrievals~\citep{retrieval-robustness, recomp}, and optimizing additional components of the retrieval pipeline~\citep{ra-dit}. 

\paragraph{Sparse Attention.} Given the quadratic computational complexity of standard attention,  {sparse attention} is chosen as a strategy to improve Transformer efficiency. Static sparse patterns include methods such as sliding window attention (SWA), dilated attention~\citep{child2019sparsetransformers, shi2021sparsebert, ding2023longnet}, and other fixed sparsity schemes. SWA mechanisms are widely adopted in many modern large language model families, including Mistral~\citep{mistral}, Phi-3~\citep{phi3}, Hymba~\citep{hymba}, Gemma-3~\citep{gemma3}, and GPT-oss~\citep{gpt-oss}.   Despite their efficiency gains, these methods typically have a limited receptive field and impose significant constraints on the flexibility of attention, particularly in attending to arbitrary token positions, thus underperforming full attention on copy-heavy tasks~\citep{swa_not_far}.

\paragraph{KV Cache Compression.} Recent research aim to improve the inference efficiency by reducing the KV cache usage during LLM decoding. SnapKV~\citep{snapkv} curates the KV cache by monitoring the accumulated attention scores and select significant tokens. ~\citet{scissorhands} propose selective dropping of low-attention KV pairs, while ~\citet{KIVI} introduce quantization techniques for compact KV representations. H\textsubscript{2}O
~\citep{H2O} provides an adaptive token eviction strategy, improving the memory usage by balancing the recent and distant information. StreamingLLM~\citep{attntion-sink} investigates and emphasize the inherent patterns of pre-trained attention, e.g. attention-sinks. By keeping the attention-sinks and local window patterns, it expands the context size of LLM on super-long inputs. PyramidKV~\citep{pyramid-kv} dynamically adjusts the KV cache consumption across different layers, which allocates more budgets for lower layers while tightening for upper layers. 
In this work, we focus on enhancing sparse attention for in-context retrieval, while KV cache compression addresses post-hoc efficiency in memory storage and reuse. Consequently, the two approaches are orthogonal: our method operates independently of KV cache compression and can be seamlessly combined with it for further efficiency gains.

\section{Problem Formulation and Overview}

This work seeks to boost the in-context recall capabilities of LLMs, with a particular emphasis on improving their performance in long-context tasks under constrained memory conditions. LLM inference generally comprises two stages: \textit{pre-filling} and \textit{decoding}. In the pre-filling stage, the model takes the user prompt as input and processes it in parallel to generate the initial hidden states.

We denote the user prompt as a sequence $x=(x_1, \dots, x_N)$ and a model with a hidden dimension $d$. For the $l$-th attention layer, we utilize the trained model weights $\mathbf{W}_Q^l, \mathbf{W}_K^l, \mathbf{W}_V^l \in \mathbb{R}^{d \times d}$ for the query, key, and value projection matrices. In the standard \textit{Full Attention} (FA) mechanism, every token $x_i$ must attend to all preceding tokens in the sequence to capture global dependencies. The query, key, and value projections for a token at position $i$ are computed as:
\begin{equation}
    [\mathbf{q}_i, \mathbf{k}_i, \mathbf{v}_i] = \mathbf{h}_i^{l-1} [\mathbf{W}_Q^l, \mathbf{W}_K^l, \mathbf{W}_V^l],
\end{equation}
where $\mathbf{h}_i^{l-1} \in \mathbb{R}^d$ is the input hidden state. The attention output $\mathbf{o}_i$ is the scaled dot-product across the entire sequence length $N$:
\begin{equation}
    \mathbf{o}_i = \text{softmax}\left(\frac{\mathbf{q}_i(\mathbf{K}^l)^\top}{\sqrt{d_k}} \right)\mathbf{V}^l.
\end{equation}

A primary challenge in deploying LLMs for long-context applications is the quadratic complexity of this operation. Because the softmax attention must be computed over all $N$ tokens for each of the $N$ positions, the computational requirements and memory consumption scale at $O(N^2)$. This memory bottleneck limits the effective context window size as the input length increases.

\subsection{LLMs with Pre-Sparsified and Post-Sparsified Attentions}
In order to balance the high performance and efficiency of LLMs, many works (e.g., ~\citet{mistral, phi3, hymba, gemma3, gpt-oss}) replace full attention in the token-mixing layer with sparse attention~\citep{SWA}. These models, referred to as \textit{pre-sparsified} LLMs, are pre-trained with sliding window attention used in all or parts of the token-mixing layers.  There is another line of work that focuses on \textit{post-sparsified} LLMs with full attention during pre-training by identifying the inherent structures of the attention layers and evict unimportant tokens in the KV cache~\citep{H2O, scissorhands, attntion-sink, lm-infinite, snapkv, Duoattention}. Both the pre-sparsified and post-sparsified LLMs can work with a manageable KV cache buffer with a limited size, which is crucial to provide affordable, efficient and high-performing services for a wide range of real-world applications.

Although many works show that the sparse characteristics of attention matrices widely exist in pre-trained LLMs~\citep{scissorhands, attntion-sink, lm-infinite}, the sparse patterns are highly context-dependent. Namely, many tokens attend only to parts of the input sequence, and on which specific tokens it focuses is highly dependent on the surrounding context, which often differs markedly across various inputs. This context-sensitive structure is a crucial feature of models' behavior that enhances in-context recall. In such settings, the ability to selectively retrieve relevant information from long or complex sequences hinges on the model’s capacity to dynamically adapt its attention based on nuanced contextual cues. Without this flexibility, the model risks overlooking key dependencies for accurate understanding and generation.

\section{\methodname: Improving Sparsified Attention via External Retriever}
In this section, we introduce our method, \methodname, designed to enhance the in-context recall capability of sparse attention mechanisms by incorporating a pre-trained external retriever.

At its core, \methodname~augments a sparse attention model with dynamically generated token positions that are useful to next-token generation. These positions are identified via a retrieval system, which is detailed in \autoref{subsec:search}.
This allows the model to directly access and integrate information from arbitrary positions in the sequence, effectively extending its receptive field far beyond the constraints of local attention windows. 
This is particularly important for tasks that involve very long inputs, where maintaining both efficiency and global contextual information remains a significant challenge.

In \autoref{subsec:attn_comp}, we demonstrate how the information from retrieval can be integrated into the attention computations. Essentially, we sparsify the attentions based on the retrieved positions and the original sparse patterns. With chunked pre-filling and limited KV recomputations during decoding, we can achieve memory- and time- efficient  sub-quadratic attention computations.

\subsection{In-Context Dense Search}
\label{subsec:search}

\begin{figure*}[t!]
    \centering
    \includegraphics[width=\linewidth]{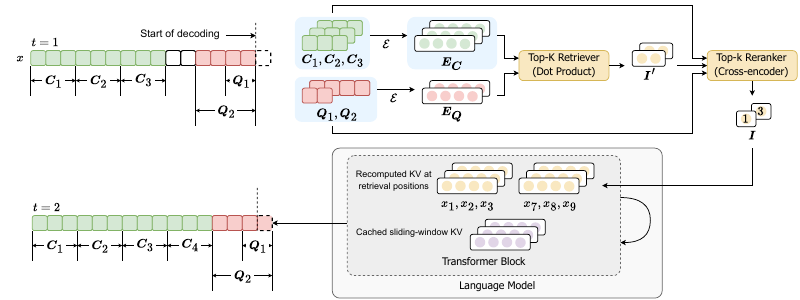}
    \caption{Illustration of one decoding step using \methodname, where $U=3$, $K=4$, and $k=2$. $\mC$ and $\mQ$ are context chunks and query chunks respectively. $\mathcal{E}$ is a Sentence-BERT encoder. Objects with a blue background can be cached in memory and used for recomputation during the generation of each token.}
    \label{fig: decode}
\end{figure*}

\methodname~incorporates an in-context search module that leverages an external retrieval system to dynamically identify the most salient context for each input token. This module should be designed so that it does not introduce high latency and memory overhead during generation. Below, we briefly describe the module.

Given an input sequence $x = (x_1, \dots, x_N)$, we first partition the sequence into fixed-size context chunks of length $U$, which constitute the candidate pool for retrieval. For each token position, we construct a \textit{query chunk} consisting of the immediate preceding tokens to provide a localized semantic representation.
During pre-filling, we retrieve the top-$k$ chunks using a bi-encoder over dense embeddings produced by a Sentence-BERT model \citep{reimers-gurevych-2019-sentence}. Bi-encoders are fast and ensures low-latency pre-filling.
During decoding, we retrieve the top-$K$ chunks $I'$ using the bi-encoder, rerank them using a cross-encoder, and select only the top-$k$ chunks, where $K > k$. Fig.~\ref{fig: decode} illustrates the idea. While cross-encoders capture token-level interactions and generally offer higher retrieval quality, this hybrid pipeline helps balance precision and speed. For techniques on further improving efficiency and more discussion about the module, see \autoref{app:retrieval} and \autoref{app:rerank}.

\subsection{Augmented Attention Computation}
\label{subsec:attn_comp}

We incorporate the retrieved results into the attention computation by unmasking the corresponding token positions, allowing each query token to attend not only to its causal local window but also to its retrieved tokens. As a result, we only need to maintain a constant-sized KV cache for the original sparse attention pattern. Moreover, the number of retrieved token positions remains fixed regardless of the sequence length. Therefore, the memory usage of sparse attention layers stays constant with respect to the overall context length, enabling better scalability to extremely long input sequences.

During inference, our method~uses token positions identified from retrieval, which can be out of the sliding window of sparsified models, whose parameters may not be conditioned on during pre-training. Therefore, we adopt continual training on the models to adapt their parameters (see \autoref{app:cont_train}). Below, we describe the detailed procedures of attention computation in pre-filling and decoding, as shown in \autoref{fig: full}.

\subsubsection{Chunkwise Pre-Filling with Retrieval Information}
\label{subsec:prefill}

When processing long sequential inputs, it is a common practice to use chunk-wise pre-filling, specifically, the input prompt is partitioned into fixed-size chunks to pre-fill the KV cache. 

Concretely, given an original input sequence of length $N$, we pad it and divide it into $n$ fixed-size chunks. During the pre-filling stage, each chunk is processed sequentially: the model computes and stores its corresponding key--value representations in the cache before proceeding to the next chunk.

During the pre-filling stage with long sequential inputs, for every query token we only need to conduct the attention operation with keys indicated by the custom attention mask matrices $\mathbf{\Lambda}$ as the following formulation:

\begin{align}
\begin{bmatrix}
\mathbf{Q}\,\,\, & \mathbf{K}\,\,\, & \mathbf{V}
\end{bmatrix}
=
\begin{bmatrix}
\mathbf{H}
\end{bmatrix}
\begin{bmatrix}
\mathbf{W}_Q & \mathbf{W}_K & \mathbf{W}_V
\end{bmatrix},
\end{align}
\begin{align}
    \mathbf{S_h^l}
    = \text{Softmax}\!\left(
        \mathbf{Q}\,
        \begin{bmatrix}
            \mathbf{K}^\top,\,
            \color{purple}\mathbf{K}_{c}^\top
        \end{bmatrix}
        + \mathbf{\Lambda}
    \right)
    \begin{bmatrix}
        \mathbf{V}\,\,\,\\
        \color{purple}\mathbf{V}_{c}
    \end{bmatrix},
\end{align}
where $\mathbf{Q}, \mathbf{K}, \mathbf{V} \in \mathbb{R}^{N\times d_{\text{head}}}$ are the query, key and value projections of one attention head, $(\textcolor{purple}{\mathbf{K}_c}, \textcolor{purple}{\mathbf{V}_c})$ are cached key--value pairs and $\mathbf{\Lambda} \in \mathbb{R}^{\bar N\times \bar N}$ is the attention mask matrix, where $\bar N = n + n_c + n_r$, $n_c$ is the size of cached KV, $n_r$ is the number of retrieved KV. In our work, the attention mask $\mathbf{\Lambda}$ consists of three components: (1) a causal mask, which guarantees no future information leakages; (2) a structured mask for SWA, which provides local information and stability; and (3) a retrieval mask, which provides relevant information from arbitrary positions in the context.

\subsubsection{Decoding with Retrieved KV Projection Recomputation} 

\begin{figure*}[t!]
    \centering
    \includegraphics[width=0.95\linewidth]{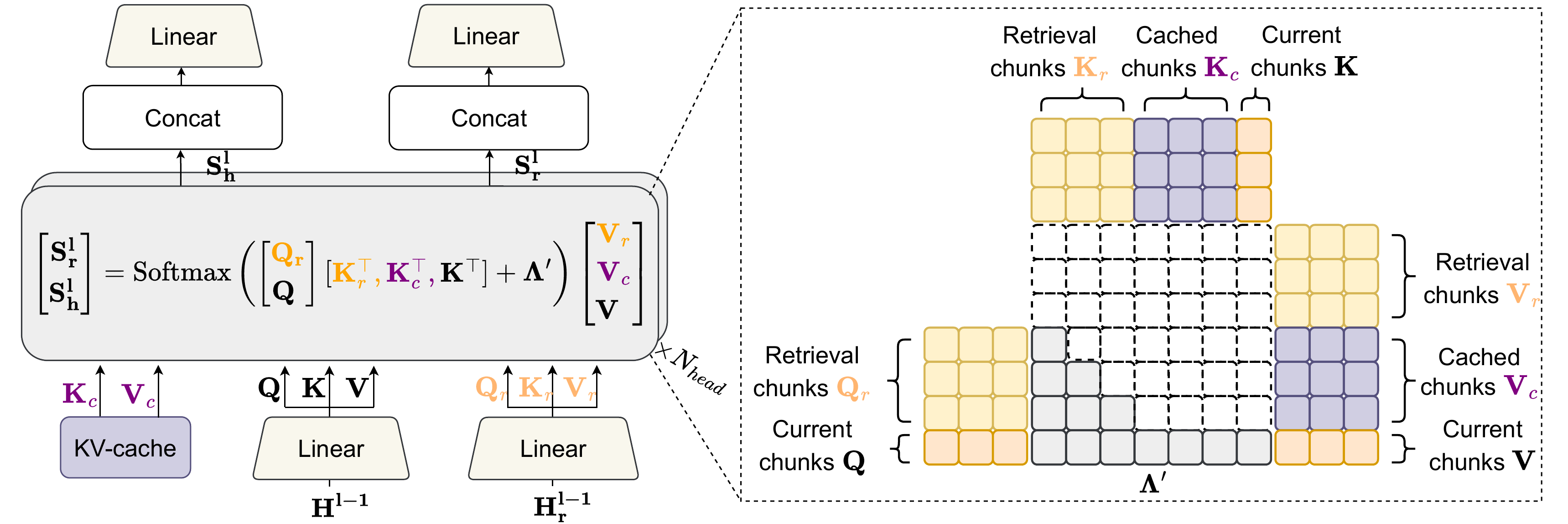}
    \caption{Illustration of the computation of \methodname~in one attention head during decoding.}
    \label{fig: full}
\end{figure*}

During each decoding step, the input sequence length is 1. The retriever identifies the corresponding position from a compact cache of original inputs, and the associated original raw input is then provided to the model as an \textit{auxiliary input}. In the subsequent forward pass, this auxiliary input participates in the attention computation synchronously with the current token and serves as the \textit{reconstructed} KV cache for its offseted position, as shown on the right hand side of ~\autoref{fig: full}. Specifically, in each layer of the attention computation, we have:

\begin{align}
\begin{bmatrix}
\color{orange}{\mathbf{Q}_r} & \color{orange}{\mathbf{K}_r} & \color{orange}{\mathbf{V}_r} \\
\mathbf{Q}\,\,\, & \mathbf{K}\,\,\, & \mathbf{V}\,\,\,
\end{bmatrix}
=
\begin{bmatrix}
\mathbf{H}_r \\
\mathbf{H}\,\,\,
\end{bmatrix}
\begin{bmatrix}
\mathbf{W}_Q & \mathbf{W}_K & \mathbf{W}_V
\end{bmatrix},
\end{align}
\begin{align}
    \begin{bmatrix}\mathbf{S_r^l} \\ \mathbf{S_h^l}\end{bmatrix} =\text{Softmax}\left(\begin{bmatrix}
\color{orange}\mathbf{Q_r}\\
\mathbf{Q}\, \,
\end{bmatrix}
\color{black}
\begin{bmatrix}
\color{orange}{\mathbf{K}_{r}^\top}\color{black},
\color{purple}\mathbf{K}_{c}^\top\color{black},
\color{black}\mathbf{K}^\top
\end{bmatrix}
+ \mathbf{\Lambda}'
\right)
\begin{bmatrix}
\color{orange}\mathbf{V}_{r} \\
\color{purple}\mathbf{V}_{c} \\
\mathbf{V}\,\,\,
\end{bmatrix},
\end{align}
where $\mathbf{\Lambda}'$ denotes the attention mask, shaped as illustrated on the right side of ~\autoref{fig: full}, and $\mathbf{H}, \mathbf{H}_r$ represent the hidden states corresponding to the main and auxiliary inputs, respectively. 

By re-computing the approximate attention over the auxiliary inputs at each layer, we efficiently while effectively reconstruct the KV cache for past inputs of arbitrary depth, enabling the model to recover long-range dependencies that would otherwise be lost due to cache truncation. 

Crucially, this resource-intensive KV re-computation is performed only once at the beginning of each decoding chunk. The resulting reconstructed KVs at each layer are then stored in a fixed-size temporary cache. As a result, all subsequent new-coming query tokens within the same chunk can directly reuse this cache without repeating the expensive re-computation, thereby ensuring both the efficiency and effectiveness of \methodname~.

This mechanism allows the retrieval-biased sparse attention to dynamically integrate relevant information that was not originally present in its limited cache, thereby enhancing both contextual coherence and overall performance in long-context scenarios.

\section{Experiments and Results}
We evaluate \methodname's performance on both synthetic and real-world long-context benchmarks, including Multi-Query Associative Recall (MQAR)~\citep{zoology_mqar}, Mechanistic Architecture Design (MAD)~\citep{madlab}, LongBench~\citep{longbench}, and Needle-in-a-Haystack (NIAH)~\citep{sled-niah}.
We follow the standard setup of models for MQAR and MAD.
For LongBench and NIAH, we experiment with a post-sparsified Qwen3-8B-Base \citep{qwen3} and the pre-sparsified Phi-3-mini-4k-instruct \citep{phi3}.

\paragraph{Retrieval Configurations.}
For LongBench and NIAH, we use all-MiniLM-L6-v2 \citep{NEURIPS2020_3f5ee243} which comprises only 22.6M parameters as the embedding model, and bge-reranker-v2-m3 \citep{li2023making, chen2024bge} which comprises 568M parameters as the reranker.
We set $U=128$, $K=64$, and $k \in \{4,8\}$.
For MQAR and MAD, since the data is not natural language, we simply use exact string matching as the retrieval method, and set $U \in \{1,2,4\}$, and $K=k=1$.
\begin{figure*}[ht!]
    \centering
    \includegraphics[width=\linewidth]{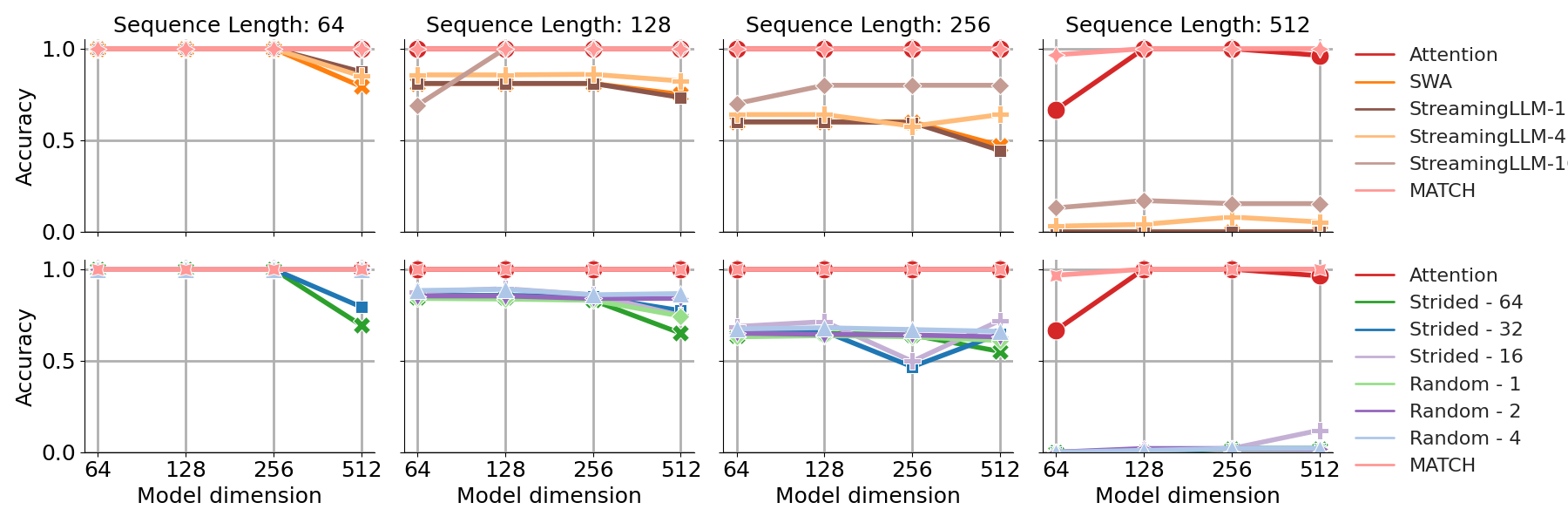}
    \caption{Results on MQAR. The top and bottom rows share identical experimental settings, differing only in the subjects being compared. Larger sequence lengths correspond to increased task difficulty. 
    }
    \label{fig: MQAR}
\end{figure*}
\begin{table}[t!]
\centering
\small
\setlength{\tabcolsep}{0.1cm}
\resizebox{1.\linewidth}{!}{
\begin{tabular}{lccccc@{}}
\toprule
\textbf{Model}      & Fuzzy ICR      & ICR            & Noisy ICR      & Avg.\\
\midrule
Hymba & 13.8          & 91.0          & 88.9          & 64.0           \\
Hymba w/ \methodname~($U=2$) & \textbf{80.3} & \textbf{99.0} & 98.0          & \textbf{92.4} \\
Hymba w/ \methodname~($U=4$) & 72.5          & 98.7          & \textbf{98.1} & 89.8          \\
\midrule
SWA   & 10.3          & 86.2          & 83.5          & 59.9          \\
SWA w/ \methodname~($U=2$)  & \textbf{72.6} & \textbf{99.0} & \textbf{96.6} & \textbf{89.4} \\
SWA w/ \methodname~($U=4$)  & 55.5          & 98.9          & 95.1          & 83.2    \\
\bottomrule
\end{tabular}
}
\caption{Performance on three in-context retrieval tasks in MAD~\citep{madlab}.
}
\label{tab: mad}
\end{table}
\subsection{Experiments on Synthetic Tasks}

We first validated the effectiveness of the \methodname~architecture on recall-intensive tasks using MQAR and MAD benchmarks. For details of the setup of synthetic experiments, see \autoref{app:mqar_details}.

\paragraph{Multi-Query Associative Recall (MQAR).}

MQAR tests the associative recall capabilities of models.
\autoref{fig: MQAR} compares \methodname~against \texttt{StreamingLLM}, \texttt{Strided}~\citep{zhang2023ho}, and \texttt{Random}, where the suffixes denote sink width, stride intervals, and random activation count respectively for the three methods. Despite requiring fewer activations, \methodname~matches the performance of full attention and significantly outperforms other methods by a huge margin. Notably, at the sequence length of 512, \methodname~maintains near-perfect accuracy whereas other sparse baselines almost degrade to random guessing.

\paragraph{Mechanistic Architecture Design (MAD).}

We conducted experiments on three synthetic in-context recall tasks proposed in the MAD suite. 
We evaluate two model types: Hymba model~\citep{hymba} and a transformer equipped with SWA.
\autoref{tab: mad} shows that our proposed \methodname~consistently boosts in-context retrieval (ICR) performance across all tasks and model variants. 
For both Hymba-based and SWA-based model, \methodname~boosts the average performance by 45\% to 50\%, these gains hold across all three ICR tasks, with \methodname~achieving the best or near-best performance in nearly all metrics.
Overall, these results highlight the effectiveness of \methodname~in context-recall–intensive scenarios.

\subsection{Experiments on LongBench}

\begin{table*}[th!]
\centering
\small
\begin{tabular}{lccccccc}
\toprule
\multirow{4}{*}{\textbf{Model}} & \multirow{4}{*}{\textbf{SWA ratio}} & \multicolumn{6}{c}{\textbf{LongBench Task Types}} \\
\cmidrule(lr){3-8}
& & \begin{tabular}{@{}c@{}}Single-Doc.\\ QA (SQ)\end{tabular} & \begin{tabular}{@{}c@{}}Multi-Doc.\\ QA (MQ)\end{tabular} & \begin{tabular}{@{}c@{}}Summ.\\(SM)\end{tabular} & \begin{tabular}{@{}c@{}}Few-shot\\Learning (FS)\end{tabular} & \begin{tabular}{@{}c@{}}Synthetic\\Tasks (ST)\end{tabular} & Average \\
\midrule 
\multicolumn{8}{c}{\textit{Post-sparsified LLM}} \\
\cmidrule(lr){1-8}
Qwen3 & \multirow{4}{*}{0.8} & 41.6 & 34.7 & 20.7 & 61.1 & 12.7 & 34.2 \\
Qwen3 w/ RAG & & 42.0  & 38.0  & 21.1  & 59.7  & 21.0  & 36.4  \\
Qwen3 w/ \methodname~($k=4$) & & \textbf{43.0} & 35.9 & 19.9 & \textbf{61.4} & \underline{24.0} & \underline{36.8} \\
Qwen3 w/ \methodname~($k=8$) & & \underline{42.1} & \underline{36.0} & 19.9 & \underline{61.2} & \textbf{26.0} & \textbf{37.0} \\
\cmidrule(lr){1-8}
Qwen3 & \multirow{4}{*}{0.5} & 43.1 & 36.5 & 21.3 & 61.7 & 36.7 & 39.9 \\
Qwen3 w/ RAG & & 44.1   & 39.2   & 20.2   & 60.8   & 42.0   & 41.2  \\
Qwen3 w/ \methodname~($k=4$) & & \textbf{44.8} & \underline{39.1} & \underline{20.7} & \underline{61.7} & \underline{50.8} & \textbf{43.4} \\
Qwen3 w/ \methodname~($k=8$) & & \underline{44.7} & {38.0}  & {20.6}  & \textbf{62.0}  & \textbf{51.8}  & \textbf{43.4}  \\
\midrule
\multicolumn{8}{c}{\textit{Pre-sparsified LLM}} \\
\cmidrule(lr){1-8}
Phi-3 & \multirow{4}{*}{1} & 24.4 & 21.9 & 20.8 & 47.0 & 4.0 & 23.6 \\
Phi-3 w/ RAG &  & 27.5  & 27.0  & 20.1  & 46.4  & 8.5  & 25.9 \\
Phi-3 w/ \methodname~($k=4$) &  & \textbf{29.6} & 26.8 & \textbf{20.8} & \underline{46.9} & 7.3 & \underline{26.3} \\
Phi-3 w/ \methodname~($k=8$) &  & \underline{28.6} & \textbf{27.1} & {20.3}  & \underline{46.9}  & \textbf{9.7} & \textbf{26.5} \\
\bottomrule
\end{tabular}
\caption{Results on LongBench. \methodname~is applied to post-sparsified and pre-sparsified LLMs. The sparsity ratio denotes the proportion of attention layers replaced with SWA. RAG and \methodname~utilize the same retrieval content.}
\label{tab: longbench}
\end{table*}

\autoref{tab: longbench} presents a comprehensive comparison between the performance of the sparsified baseline models with and without \methodname~and the best-performing RAG results (see \autoref{app:rag}) on the diverse downstream tasks of on LongBench.
First, across all sparsity levels and model types, \methodname~consistently improves overall average performance. The most pronounced improvements occur on synthetic tasks (up to +15.1), demonstrating that \methodname~effectively recovers task-specific reasoning ability that is typically degraded by sparsification. Second, we also compare \methodname~against retrieval-augmented generation (RAG) that uses the same retrieved results. \autoref{tab: longbench} reports  The results show that \methodname~is mostly on par with or better than RAG, underpinning how \methodname~can be a better alternative to incorporate retrieved contexts than RAG.

When $k=4$, \methodname~performs equally well with $k=8$. The ablation studies presented in \autoref{subsec:retrieval_ablation} show that performance drastically deteriorates under random retrieval. This shows that the retriever has high recall.

\subsection{Experiments on NIAH}
\begin{table*}[t!]
\centering
\small
\setlength{\tabcolsep}{0.15cm}
\begin{tabular}{lcccccccccccc}
\toprule
\multirow{3}{*}{{\textbf{Model}}} 
& \multicolumn{2}{c}{{8K}} 
& \multicolumn{2}{c}{{16K}} 
& \multicolumn{2}{c}{{32K}} 
& \multicolumn{2}{c}{{64K}} 
& \multicolumn{2}{c}{{128K}} 
& \multirow{3}{*}{{Avg.}} 
& \multirowcell{2}[-1ex]{Avg. \\ $\leq$32K} \\
\cmidrule(lr){2-3} \cmidrule(lr){4-5} \cmidrule(lr){6-7} \cmidrule(lr){8-9}  \cmidrule(lr){10-11}
& {Single} & {Multi} 
& {Single} & {Multi} 
& {Single} & {Multi} 
& {Single} & {Multi} 
& {Single} & {Multi} 
& & \\ 
\midrule 
\multicolumn{13}{c}{\textit{Post-sparsified LLM}} \\
\cmidrule(lr){1-13}
{Qwen3}    & 87.0 & 91.9 & 80.0 & 84.1 & 72.7 & 58.0 & 90.0 & 24.4 & 47.0 & 13.6 & 60.3 & 78.7 \\
Qwen3 w/ \methodname~($k=4$)     & \textbf{100.0} & \underline{92.4} & \textbf{99.7} & 81.5 & \textbf{93.0} & \textbf{59.7} & \underline{84.7} & \textbf{29.0} & \textbf{80.7} & \textbf{21.0} & \textbf{69.8} & \textbf{85.2} \\
Qwen3 w/ \methodname~($k=8$)     & \underline{99.0} & \textbf{93.0} & \underline{99.0} & \underline{82.3} & \underline{91.3} & 55.4 & 77.3 & \underline{26.6} & \underline{71.0} & \underline{19.9} & \underline{67.5} & \underline{84.2} \\
\midrule
\multicolumn{13}{c}{\textit{Pre-sparsified LLM}} \\
\cmidrule(lr){1-13}
{Phi-3}     & 24.3 & 18.2 & 14.3 & 8.2 & 3.7 & 4.2 & -- & -- & -- & -- & 11.7 & 11.7 \\
{Phi-3} w/ \methodname~($k=4$)    & \underline{78.0} & \underline{43.6} & \underline{75.3} & \underline{40.5} & \underline{75.3} & \textbf{36.9} & -- & -- & -- & -- & \underline{53.8} & \underline{53.8} \\
{Phi-3} w/ \methodname~($k=8$)    & \textbf{80.0} & \textbf{47.5} & \textbf{78.7} & \textbf{45.0} & \textbf{77.0} & \underline{36.7} & -- & -- & -- & -- & \textbf{56.4} & \textbf{56.4} \\
\bottomrule
\end{tabular}
\caption{
{Results on NIAH across context lengths.} `--' denotes out-of-memory runs.
}
\label{tab: niah}
\end{table*}

\autoref{tab: niah} presents the performance comparison of \methodname~applied to both post-sparsified and pre-sparsified LLMs on the NIAH benchmark under varying context lengths (8K--128K). The “Single” and “Multi” columns respectively report single-instance and multi-instance retrieval accuracies. Across all context settings, our methods consistently outperform the corresponding sparsified baselines (Qwen3 and Phi-3). The improvements are particularly pronounced in long-context scenarios (32K--128K), demonstrating that \methodname~effectively mitigates degradation in retrieval performance as sequence length increases. When excluding the out-of-memory runs of the 64K and 128K setups, our approach yields higher average scores, confirming its robustness and generalization across different sparsity configurations.

\section{Analysis and Discussion}
\subsection{Ablation Studies}
\label{subsec:retrieval_ablation}
\begin{table}[htb!]
\centering
\small
\setlength{\tabcolsep}{0.19cm}
\begin{tabular}{lccccccc}
\toprule
\multirow{3}{*}{\textbf{Method}} & \multicolumn{6}{c}{\textbf{LongBench Task Types}} \\
\cmidrule(lr){2-7}
 & SQ & MQ & SM & FS & ST & Avg. \\
\midrule
Base & 41.6 & 34.7 & 20.7 & 61.1 & 12.7 & 34.1 \\
\, w/ Random & 40.0 & 34.7 & 19.6 & 60.1 &  9.5 & 32.8 \\
\, w/ \methodname & \textbf{42.1} & \textbf{36.0} & 19.9 & \textbf{61.2} & \textbf{26.0} & \textbf{37.0} \\
\bottomrule
\end{tabular}
\caption{Results of ablation of retrieval.}
\label{tab:retrieval_ablation}
\end{table}

To verify that \methodname’s gains are attributed to the introduction of the retrieved context rather than the mere increased exposure of attention, we replace the retriever with a random index generator while keeping all other hyperparameters identical. As shown in \autoref{tab:retrieval_ablation}, using Qwen3 with 80\% SWA sparsity and $k=8$, the random-retrieval model consistently underperformed both full \methodname~and even the pure sparse-attention baseline, indicating that naïvely adding extra KV pairs is ineffective and even harmful. This confirms the necessity of \methodname’s carefully designed retrieval pipeline.

\begin{figure*}[th!]
     \centering
    \includegraphics[width=1\linewidth]{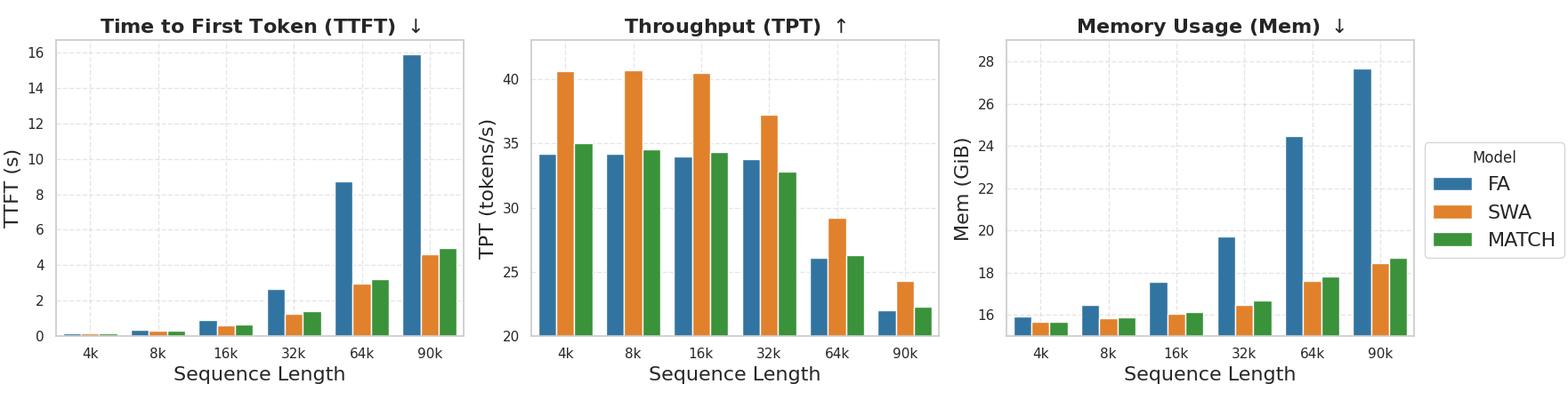}
    \caption{Performance comparison of attention mechanisms across sequence lengths: (a) Time to First Token (TTFT), (b) Throughput (Tokens Per Second), and (c) Memory consumption. Our method achieves competitive latency and throughput while maintaining significantly better memory efficiency than FA and comparable performance to SWA.}
    \label{fig:speed}
\end{figure*}

\begin{table*}[t!]
\centering
\small
\setlength{\tabcolsep}{0.15cm}
\begin{tabular}{lcccccccc}
\toprule

\multirow{4}{*}{\textbf{Method}} & \multicolumn{6}{c}{\textbf{LongBench Task Types}} & \multirow{4}{*}{\textbf{TTFT (s)} $\downarrow$} & \multirow{4}{*}{\begin{tabular}{@{}c@{}}\textbf{Decoding}\\\textbf{Memory (GiB)}\end{tabular} $\downarrow$ } \\
\cmidrule{2-7}
& \begin{tabular}{@{}c@{}}Single-\\Doc. QA\end{tabular} $\uparrow$ & \begin{tabular}{@{}c@{}}Multi-\\Doc. QA\end{tabular} $\uparrow$ & Summ. $\uparrow$ & \begin{tabular}{@{}c@{}}Few-shot\\Learning\end{tabular} $\uparrow$ & \begin{tabular}{@{}c@{}}Synthetic\\Tasks\end{tabular} $\uparrow$ & Avg. $\uparrow$ & & \\
\midrule
StreamingLLM & 42.5 & 34.2 & 24.7 & 59.3 & 33.7 & 38.9 & 1.8 & 19.8 \\
FlexPrefill & 43.3 & 37.9 & 24.8 & 58.5 & 26.5 & 38.2 & 1.6 & 19.7 \\
\methodname & \textbf{44.7} & \textbf{38.0} & 20.6 & \textbf{62.0} & \textbf{51.8} & \textbf{43.4} & \textbf{1.4} & \textbf{16.7} \\
\bottomrule
\end{tabular}
\caption{Results on LongBench, time to first token (TTFT), and memory required for decoding by different attention sparsification methods.
}
\label{tab:qwen3_compare}
\end{table*}

\subsection{Comparison with KV-cache compression techniques}

\begin{table*}[t!]
\centering
\small
\begin{tabular}{lcccccccccc}
\toprule
\multirow{3}{*}{{\textbf{Method}}} 
& \multicolumn{2}{c}{{8K}} 
& \multicolumn{2}{c}{{16K}} 
& \multicolumn{2}{c}{{32K}} 
& \multicolumn{2}{c}{{64K}} 
& \multirow{3}{*}{{Average}} \\
\cmidrule(lr){2-3} \cmidrule(lr){4-5} \cmidrule(lr){6-7} \cmidrule(lr){8-9}
& {Single} & {Multi} 
& {Single} & {Multi} 
& {Single} & {Multi} 
& {Single} & {Multi} \\ 
\midrule
StreamingLLM & 70.7 & 65.4 & 50.0 & 38.3 & 49.7 & 22.0 & 60.7 & 14.8 & 43.6 \\
FlexPrefill & 93.7 & 76.9 & 96.0 & 74.4 & 98.3 & 66.8 & 84.7 & 39.7 & 75.2 \\
\methodname & \textbf{100.0} & \textbf{92.4} & \textbf{99.7} & \textbf{81.5} & 93.0 & 59.7 & \textbf{84.7} & 29.0 & \textbf{76.4} \\
\bottomrule
\end{tabular}
\caption{
{Results on NIAH by different attention sparsification methods.}
}
\label{tab: niah_compare}
\end{table*}

We further compare \methodname~against two prominent sparse attention baselines StreamingLLM~\citep{xiao2024efficient} and FlexPrefill~\citep{lai2025flexprefill}  (for details, see \autoref{app:baselines}). 
We applied these two methods directly to the standard Qwen3 model, adhering to the hyperparameter settings recommended in their original papers to ensure optimal performance. \Autorefs{tab:qwen3_compare} and \ref{tab: niah_compare} present the results on LongBench and NIAH respectively. We also report the Time-to-First-Token (TTFT) latency and GPU memory footprint during a typical LongBench sample inference for all three approaches. 

We apply \methodname~to the 50\%-sparsified Qwen3 with $k=8$, and it~achieves the highest overall average score of 43.4 on LongBench, outperforming both StreamingLLM and FlexPrefill. It delivers clear gains in most of the task types except summarization. In addition, \methodname~records the lowest latency and decoding memory footprint at 32K context length, highlighting its time and space efficiency advantage. We also observe that \methodname~with $k=4$ mostly outperforms the two other methods on NIAH. 

Overall, the results indicate that \methodname~enhances sparsified LLMs by restoring their representational and reasoning capabilities, leading to systematic performance gains without compromising sparsity efficiency. This highlights the potential of \methodname~as a general plug-in framework for boosting the fidelity and reliability of sparse long-context models.

\subsection{Efficiency Analysis}
To better understand the advantages brought about by \methodname~in handling long sequences, we conduct an additional experimental study to evaluate its computational efficiency relative to the backbone models it augments. In \autoref{fig:speed}, we present ablations using Qwen3 with $k=8$, comparing the base models with those enhanced by our method to quantify the incremental overhead introduced across three metrics—Time to First Token, Throughput, and Memory Consumption—over sequence lengths ranging from 4K to 90K tokens. 
While the \methodname~module introduces a marginal increase in each computational factor, it remains lightweight, and the overall overhead compared to SWA is negligible, particularly for long sequences. Moreover, the augmented system achieves substantial efficiency gains compared to transformer models with full attention on our tasks.

Specifically, \methodname~achieves a TTFT of 4.98s at 90K tokens, which is a 68\% reduction from FA (15.88s) and closely comparable to SWA (4.62s). Furthermore, throughput remains stable across increasing sequence lengths, closely tracking FA’s scaling behavior, while memory usage is reduced by 32\% relative to FA (18.72 vs. 27.67 GiB). Overall, the method maintains a favorable balance among latency, throughput, and memory consumption, demonstrating its practicality for scalable long-context inference without compromising performance fidelity.

\section{Conclusion}
In this work, we propose \methodname, a lightweight retrieval-based knowledge integration mechanism that enhances the ability of sparsified LLMs to leverage example-specific context. \methodname~introduces a novel approach that treats the input context as a dynamic datastore for retrieval, integrating the retrieved information with the input during both training and inference. This design enables models to utilize contextual information more effectively while overcoming the limitations imposed by local attention or data-independent sparse patterns. Across a variety of synthetic and real-world data sets, sparse-attention LLMs augmented with \methodname~achieve substantial performance improvements over their base counterparts, highlighting its effectiveness as a general and broadly applicable framework.
\section{Limitations}
In this work, we focus on in-context retrieval tasks. We believe that a concentrated examination of these tasks allows us to delve deeper into the nuances and intricacies involved, thereby providing more insightful and meaningful findings. However, examining a more diverse range of tasks may help us to further and more broadly assess the effectiveness of our method. Furthermore, our method is designed to improve the context-copy ability of sparse attentions, and it operates without optimizing layer selection. Investing more time and resources into these aspects may further strengthen the method. We will leave these for future work.
\bibliography{custom}

\appendix

\section{Details of Retrieval}
\label{app:retrieval}
To reduce the latency of pre-filling on long prompts, we perform retrieval not at every token position but only for the last $m$ token positions, and at regular intervals so that the same retrieved chunks are shared within each interval of $U$ tokens. Concretely, $x$ is chunked as:
\begin{align}
      (x_1, \dots, x_N) & \xrightarrow{\textup{chunk}} (\mC_1, \dots, \mC_T) \nonumber\\
      \textup{where}~ \mC_i & = (x_{(i-1)U + 1}, \dots, x_{iU}), \nonumber\\
      T &= \lfloor N / U \rfloor. \nonumber
\end{align}
We then perform exact searches over the dense embeddings of chunks using a bi-encoder, which can be efficiently parallelized via matrices:
\begin{align}
\label{eq:emb}
\mathbf{E}_i & = \mathcal{E}(\mC_i) \quad i\in[1,T], \\
\label{eq:dot}
{\mathbf{S}} &= ({\mathbf{E}_{\lfloor\frac{N-m}{U}\rfloor:T}}\mathbf{E}^\top) + \mathbf{M} \\
    & \textup{where}~ \mathbf{M}_{ij} =
    \begin{cases}
    0, & i > j \\
    -\infty & \textup{otherwise}
    \end{cases}. \nonumber
\end{align}
Finally, the top-$k$ chunks per token position $I' = \{\vs_1, ..., \vs_N\}$ are obtained by
\begin{align}
I'_i &= 
    \begin{cases}
    \operatorname{arg\,topk}_{j}({\mathbf{S}}_{\tau j}) & \tau \in [N-m+1, N]\\
    \varnothing & \textup{otherwise}
    \end{cases}
\label{eq:argtopk}
\end{align}
where $x_i \in \mC_\tau$ and $\mathcal{E}$ is the Sentence-BERT encoder.

The above is performed once per input sequence, incurring one-off computational costs of $O(TE)$ for \eqref{eq:emb} where $E$ is the complexity of a forward pass of the Sentence-BERT model, and $O(T^2D)$ for \eqref{eq:dot} where $D$ is the embedding dimension.

During decoding, we perform multi-query search, allowing retrieval at different granularities. We denote $P$ as the lengths of the set of queries. For example, given $P=\{64,128\}, K=64, k=8$, two queries are formed by taking the preceding 64 and 128 tokens respectively. Then, the bi-encoder would retrieve $\lfloor K/\lvert P \rvert\rfloor = 64/2 = 32$ chunks, and $\lfloor k/2 \rfloor = 8 / 2 = 4$ chunks will be selected for each query after reranking. We further discuss the design choices of reranking in \autoref{app:rerank}. As in pre-filling, retrieval is also performed at intervals to minimize latency.

\section{Experimental Details}

\subsection{Adapting LLMs via Continual Training}
\label{app:cont_train}
Weights of LLMs pre-trained with full attention or sliding window attention ~\citep{mistral, llama3, qwen2.5, phi3, gemma3, qwen3} are conditioned to operate under the respective attention setups.
A direct replacement of original attention layers by \methodname's~ones could introduce significant discrepancies between training and inference.
Consequently, we propose adaptations via continual pretraining. 

For LLMs with both post-sparsified and pre-sparsified SWA, we conduct a two-stage continual pre-training. For the first stage,  we sampled 25B data from Cosmopedia and 25B from Fineweb-edu. We also collect data from several curated data sets, including 14B data extracted from the ProLong dataset, a long-QA retrieval dataset containing contexts, questions, and answer pairs. Long paragraphs of gathered results with corresponding answers were generated by an OpenAI chat agent. The second stage utilizes 1.5B Prolong short-mix data, 1.5B NIAH-like synthetic data, and 1B Narriative QA dataset.

\subsection{Tasks of LongBench}
The tasks of LongBench~\citep{longbench} we evaluated on are listed in \autoref{tab: longbench_data}.
\label{sec:dataset}

\begin{table*}[h!]
\centering
\small
\label{tab:longbench_datasets}
\begin{tabular}{cccc}
\toprule
\textbf{Dataset Name} & \textbf{Context Type} & \textbf{Avg. Length (Tokens)} & \textbf{Evaluation Metrics} \\
\midrule
2WikiMQA & Multi-document & 2.5K & F1\\
DuReader & Single-document & 1.6K & Rouge-L \\
Gov\_Report & Single-document & 9.4K & Rouge-L \\
HotpotQA & Multi-document & 10.8K & F1 \\
LSHT & Single-document & 13.5K & Accuracy \\
Multi-News & Multi-document & 11.8K & Rouge-L \\
MultiFieldQA-en & Multi-document & 4.2K & F1 \\
MultiFieldQA-zh & Multi-document & 4.3K & F1 \\
Musique & Multi-document & 11.8K & F1 \\
NarrativeQA & Single-document & 20.6K & F1 \\
Passage Count & Single-document & 9.1K & Accuracy \\
Passage Retrieval-en & Multi-document & 10.8K & Accuracy \\
Passage Retrieval-zh & Multi-document & 10.8K & Accuracy \\
Qasper & Single-document & 4.9K & F1 \\
QMSum & Multi-document & 9.8K & Rouge-L \\
Samsum & Dialogue & 0.6K & Rouge-L \\
TREC & Single-document & 5.0K & Accuracy \\
TriviaQA & Multi-document & 7.8K & F1 \\
VCSum & Dialogue & 10.3K & Rouge-L \\
\bottomrule
\end{tabular}
\caption{Tasks from LongBench on which we evaluate.}
\label{tab: longbench_data}
\end{table*}

\subsection{Details of Experiment on Synthetic Tasks}\label{app:mqar_details}

\subsubsection{MQAR Setup}
In our MQAR experiment setting, all models were implemented as 2-layer sequence mixers, trained on 100K samples, and evaluated on a hold out test set of 3K samples, following standard MQAR experiment settings. Sequence lengths ranged from 64 to 512, containing 4 to 64 recall pairs, respectively. For each experiment, we performed a sweep over four learning rates from $10^{-4}$ to $10^{-2}$ and reported the best performance. All Sparse Attention models shares a window size of 32.
We averaged the performance across three experimental runs with different random seeds.

\subsubsection{MAD Setup}
For each task in MAD, we report the accuracy on a held-out test set, where a prediction is considered correct only if the entire output sequence is exactly matched.
All models are trained from scratch on the target task using a 4-layer architecture with a vocabulary size of 16 for ICR and FuzzyICR and 32 for NoisyICR, respectively. Each model has a hidden size of 128, and for models augmented with \methodname, we tried two variants of context chunk size of 2 and 4.

\subsection{Configurations of Sparse Attention Baselines}
\label{app:baselines}
We follow the recommended configurations in the original papers of StreamingLLM and FlexPrefill. For StreamingLLM, we set \texttt{global\_window} to be 1024 and \texttt{local\_window} to be 2048. For FlexPrefill, we set \texttt{block\_size} to be 128, $\gamma$ to be 0.9, $\tau$ to be 0.1, and \texttt{min\_budget} to be 512.

\subsection{Comparing with Retrieval-Augmented Generation}
\label{app:rag}
We have tried different ways of performing RAG. This includes appending the retrieved chunks to the end of the input sequence and replacing parts of the texts in the sliding window with the chunks.
We find that appending the chunks to the end yields substantially worse performance.
We report the best-scoring RAG results on LongBench.

\section{Additional Results}

In the following subsections, we present more experimental results evaluated on LongBench for understanding the effects of different components in \methodname. For brevity, we denote single-doc. QA, multi-doc. QA, summarization, few-shot learning, and synthetic tasks as SQ, MQ, SM, FS, and ST respectively in the column headers of the tables below.

\subsection{Effect of Reranking}
\label{app:rerank}

\begin{table}[th!]
\centering
\small
\setlength{\tabcolsep}{0.19cm}
\begin{tabular}{lccccccc}
\toprule
\multirow{3}{*}{$k$} & \multirow{3}{*}{\textbf{Rerank}} & \multicolumn{6}{c}{\textbf{LongBench Task Types}} \\
\cmidrule(lr){3-8}
& & SQ & MQ & SM & FS & ST & Avg. \\
\midrule
$8$ & Yes & \textbf{42.1} & \textbf{36.0} & \textbf{19.9} & 61.2 & \textbf{26.0} & \textbf{37.0} \\
$8$ & No  & 41.1 & 34.8 & 19.5 & \textbf{61.4} & 21.7 & 35.7 \\
\midrule
$4$ & Yes & \textbf{43.0} & \textbf{35.9} & \textbf{20.0} & \textbf{61.4} & \textbf{24.0} & \textbf{36.9} \\
$4$ & No  & 40.6 & 35.3 & 19.9 & 60.5 & 22.7 & 35.8 \\
\bottomrule
\end{tabular}
\caption{Results of \methodname~with and without the reranking step in the retriever.}
\label{tab:rerank_ablation}
\end{table}

\begin{table*}[t]
\centering
\small
\begin{tabular}
{lcccccc}
\toprule
\multirow{3}{*}{$(U, P, k, m)$} & \multicolumn{6}{c}{\textbf{LongBench Task Types}} \\
\cmidrule(lr){2-7}
& SQ & MQ & SM & FS & ST & Avg. \\
\midrule
$(128,\{64,128\}, 8, 1000)$         & 42.1 & 36.0 & 19.9 & 61.2 & 26.0 & 37.0 \\
$(128,\{64,128\}, 8, -4096)$        & 41.7 & 35.8 & 19.9 & 60.9 & 27.3 & 37.1 \\
$(128,\{128\}, 4, 1000)$            & 42.3 & 35.3 & 19.9 & 61.0 & 22.7 & 36.2 \\
$(128,\{32,64,96,128\}, 8, 1000)$   & 42.4 & 36.9 & 19.6 & 60.2 & 24.5 & 36.7 \\
$(64,\{64\}, 8, 1)$                 & 42.0 & 35.3 & 19.4 & 61.4 & 22.2 & 36.1 \\
$(64,\{64\}, 8, 1000)$              & 41.8 & 35.6 & 19.3 & 61.4 & 23.7 & 36.4 \\
\bottomrule
\end{tabular}
\caption{Results of \methodname~with different configurations of hyperparameters}
\label{tab:hyp_sen}
\end{table*}

\autoref{tab:rerank_ablation} shows the results of applying \methodname~to the Qwen3 model with and without reranking during decoding. It is seen that the models with reranking almost always outperform those that do not. Nevertheless, as the reranker adopts a cross-encoder architecture, it is costly in terms of memory and time to perform pairwise scoring during pre-filling when all query chunks in the input sequence have to undergo retrieval. For instance, if we set the hyperparameters to $P = \{64,128\}$, $U=K=128$, and $m=3000$ (refer to \autoref{app:retrieval} for details), it would take up to 4 seconds for pre-filling. While it is possible to also add reranking during pre-filling, we wish to make our framework practical across hyperparameters settings. Therefore, we choose to rely solely on the efficient matrix-based bi-encoder (equations \ref{eq:emb}, \ref{eq:dot}, and \ref{eq:argtopk}) during pre-filling for applicability of different hyperparameters and simplicity.

\subsection{Sensitivity to Hyperparameters}

\autoref{tab:hyp_sen} presents the results of \methodname~under different hyperparameter configurations. Several observations can be drawn. First, performance remains relatively stable across configurations, indicating that \methodname~is not overly sensitive to hyperparameter choices. Second, all configurations yield substantial improvements over the baseline (see \autoref{tab: longbench}).

From our experiments, we find that increasing $m$ (from 0) generally yields better performances but only very marginally when $m$ is large enough. In \autoref{tab:hyp_sen}, $m=-4096$ means retrieval is performed from 4096th position onward, which generally covers more tokens than when $m=1000$ for LongBench. From this empirical insight, while there may be opportunities to optimize over m, we do not expect significant benefits by making it dynamic. We set $m$ to be 1000 for simplicity in this work. We also find that a larger chunk size (128 over 64) is also often better. Moreover, retrieval using multiple queries indeed can help increase the granularity and consequently boost performances, but it can bring adverse effects when too many queries are used.

\subsection{Detailed MQAR Results}
The detailed results for plotting \autoref{fig: MQAR} are shown in \autoref{tab:mqar_detail}.

\begin{table*}[th!]
\centering
\small
\begin{tabular}{lcccc}
\toprule
\multirow{3}{*}{\textbf{Method}} & \multicolumn{4}{c}{\textbf{Model Dimension ($d$)}} \\
\cmidrule(lr){2-5}
 & {64} & {128} & {256} & {512} \\
\midrule
\multicolumn{5}{c}{\textit{Sequence Length = 64}} \\
\midrule
Attention & 1.00 (0.00) & 1.00 (0.00) & 1.00 (0.00) & 1.00 (0.00) \\
SWA & 1.00 (0.00) & 1.00 (0.00) & 1.00 (0.00) & 0.79 (0.36) \\
StreamingLLM-1 & 1.00 (0.00) & 1.00 (0.00) & 1.00 (0.00) & 0.87 (0.11) \\
StreamingLLM-4 & 1.00 (0.00) & 1.00 (0.00) & 1.00 (0.00) & 0.85 (0.26) \\
StreamingLLM-16 & 1.00 (0.00) & 1.00 (0.00) & 1.00 (0.00) & 1.00 (0.00) \\
Strided - 64 & 1.00 (0.00) & 1.00 (0.00) & 1.00 (0.00) & 0.69 (0.31) \\
Strided - 32 & 1.00 (0.00) & 1.00 (0.00) & 1.00 (0.00) & 0.79 (0.36) \\
Strided - 16 & 1.00 (0.00) & 1.00 (0.00) & 1.00 (0.00) & 1.00 (0.00) \\
Random - 1 & 1.00 (0.00) & 1.00 (0.00) & 1.00 (0.00) & 1.00 (0.00) \\
Random - 2 & 1.00 (0.00) & 1.00 (0.00) & 1.00 (0.00) & 1.00 (0.00) \\
Random - 4 & 1.00 (0.00) & 1.00 (0.00) & 1.00 (0.00) & 1.00 (0.00) \\
\methodname~ & 1.00 (0.00) & 1.00 (0.00) & 1.00 (0.00) & 1.00 (0.00) \\
\midrule
\multicolumn{5}{c}{\textit{Sequence Length = 128}} \\
\midrule
Attention & 1.00 (0.00) & 1.00 (0.00) & 1.00 (0.00) & 1.00 (0.00) \\
SWA & 0.81 (0.00) & 0.81 (0.00) & 0.81 (0.00) & 0.75 (0.10) \\
StreamingLLM-1 & 0.81 (0.00) & 0.81 (0.00) & 0.81 (0.00) & 0.73 (0.06) \\
StreamingLLM-4 & 0.86 (0.01) & 0.86 (0.01) & 0.86 (0.00) & 0.82 (0.03) \\
StreamingLLM-16 & 0.69 (0.32) & 1.00 (0.00) & 1.00 (0.00) & 1.00 (0.00) \\
Strided - 64 & 0.85 (0.01) & 0.85 (0.01) & 0.83 (0.00) & 0.65 (0.00) \\
Strided - 32 & 0.86 (0.01) & 0.86 (0.01) & 0.85 (0.00) & 0.77 (0.09) \\
Strided - 16 & 0.88 (0.01) & 0.89 (0.01) & 0.86 (0.00) & 0.75 (0.06) \\
Random - 1 & 0.84 (0.00) & 0.84 (0.01) & 0.83 (0.00) & 0.74 (0.12) \\
Random - 2 & 0.85 (0.01) & 0.85 (0.01) & 0.84 (0.00) & 0.84 (0.00) \\
Random - 4 & 0.88 (0.01) & 0.89 (0.00) & 0.86 (0.02) & 0.87 (0.01) \\
\methodname~ & 1.00 (0.00) & 1.00 (0.00) & 1.00 (0.00) & 1.00 (0.00) \\
\midrule
\multicolumn{5}{c}{\textit{Sequence Length = 256}} \\
\midrule
Attention & 1.00 (0.00) & 1.00 (0.00) & 1.00 (0.00) & 1.00 (0.00) \\
SWA & 0.60 (0.00) & 0.60 (0.00) & 0.60 (0.00) & 0.47 (0.12) \\
StreamingLLM-1 & 0.60 (0.00) & 0.60 (0.00) & 0.60 (0.01) & 0.44 (0.05) \\
StreamingLLM-4 & 0.64 (0.00) & 0.64 (0.00) & 0.58 (0.11) & 0.64 (0.00) \\
StreamingLLM-16 & 0.70 (0.00) & 0.80 (0.00) & 0.80 (0.00) & 0.80 (0.00) \\
Strided - 64 & 0.64 (0.01) & 0.65 (0.00) & 0.64 (0.02) & 0.55 (0.14) \\
Strided - 32 & 0.65 (0.00) & 0.66 (0.00) & 0.47 (0.33) & 0.65 (0.01) \\
Strided - 16 & 0.69 (0.01) & 0.71 (0.01) & 0.50 (0.36) & 0.72 (0.02) \\
Random - 1 & 0.63 (0.00) & 0.64 (0.01) & 0.63 (0.00) & 0.61 (0.03) \\
Random - 2 & 0.65 (0.00) & 0.64 (0.01) & 0.64 (0.00) & 0.63 (0.00) \\
Random - 4 & 0.67 (0.01) & 0.68 (0.00) & 0.67 (0.00) & 0.66 (0.02) \\
\methodname~ & 1.00 (0.00) & 1.00 (0.00) & 1.00 (0.00) & 1.00 (0.00) \\
\midrule
\multicolumn{5}{c}{\textit{Sequence Length = 512}} \\
\midrule
Attention & 0.67 (0.58) & 1.00 (0.00) & 1.00 (0.00) & 0.96 (0.06) \\
SWA & 0.00 (0.00) & 0.00 (0.00) & 0.00 (0.00) & 0.00 (0.00) \\
StreamingLLM-1 & 0.00 (0.00) & 0.00 (0.00) & 0.00 (0.00) & 0.00 (0.00) \\
StreamingLLM-4 & 0.03 (0.00) & 0.04 (0.03) & 0.08 (0.00) & 0.05 (0.04) \\
StreamingLLM-16 & 0.13 (0.00) & 0.17 (0.00) & 0.15 (0.04) & 0.15 (0.04) \\
Strided - 64 & 0.00 (0.00) & 0.00 (0.00) & 0.02 (0.00) & 0.02 (0.00) \\
Strided - 32 & 0.00 (0.00) & 0.00 (0.01) & 0.02 (0.00) & 0.02 (0.00) \\
Strided - 16 & 0.00 (0.00) & 0.01 (0.02) & 0.02 (0.00) & 0.12 (0.09) \\
Random - 1 & 0.00 (0.00) & 0.01 (0.00) & 0.02 (0.00) & 0.02 (0.01) \\
Random - 2 & 0.00 (0.00) & 0.02 (0.00) & 0.02 (0.00) & 0.02 (0.01) \\
Random - 4 & 0.00 (0.00) & 0.01 (0.00) & 0.02 (0.00) & 0.02 (0.00) \\
\methodname~ & 0.97 (0.06) & 1.00 (0.00) & 1.00 (0.00) & 1.00 (0.00) \\
\bottomrule
\end{tabular}
\caption{Detailed MQAR performance comparison (accuracy) across different sequence lengths and model dimensions. The scores are the average of three runs with different seeds. The numbers in brackets are the standard deviations.}
\label{tab:mqar_detail}
\end{table*}

\end{document}